\documentclass{article}
\usepackage{spconf,amsmath,graphicx,hyperref}
\usepackage{hyphenat}
\usepackage{amssymb}
\usepackage{booktabs}
\usepackage{multirow} 

\title{PRISM: Progressive Rain removal with Integrated State-space Modeling}
\name{Pengze Xue$^{1}$, Shanwen Wang$^{1}$, Fei Zhou$^{2}$, Yan Cui$^{3}$, Xin Sun$^{1}$\sthanks{This work is supported by the Science and Technology Development Fund, Macao SAR No.0006/2024/RIA1 and National Natural Science Foundation of China No.61971388.}}
\address{$^{1}$Faculty of Data Science, City University of Macau, SAR Macao, China\\
$^{2}$College of Oceanography and Space Informatics, China University of Petroleum (East China), China\\
$^{3}$Zhuhai 4Dage Network Technology, China\\}
\begin{document}
\ninept
\maketitle
\begingroup
\renewcommand\thefootnote{}
\footnotetext{\textcopyright~2025 IEEE. Personal use of this material is permitted. Permission from IEEE must be obtained for all other uses, in any current or future media, including reprinting/republishing this material for advertising or promotional purposes, creating new collective works, for resale or redistribution to servers or lists, or reuse of any copyrighted component of this work in other works.}
\addtocounter{footnote}{-1}
\endgroup
\begin{abstract}
Image deraining is an essential vision technique that removes rain streaks and water droplets, enhancing clarity for critical vision tasks like autonomous driving. However, current single-scale models struggle to the fine-grained recovery with global consistency. To address this challenge, we propose Progressive Rain removal with Integrated State-space Modeling (PRISM), a progressive three-stage framework: ‌Coarse Extraction Network (CENet)‌, ‌Frequency Fusion Network (SFNet)‌ and ‌Refine Network (RNet)‌‌. Specifically, CENet and SFNet utilize a novel Hybrid Attention UNet (HA-UNet) for multi-scale feature aggregation by combining channel attention with windowed spatial transformers. Moreover, we propose Hybrid Domain Mamba (HDMamba) for SFNet to jointly model spatial semantics and wavelet domain characteristics. Finally, RNet recovers the fine-grained structures via original-resolution subnetwork. Our model learns high-frequency rain characteristics while preserving structural details and maintaining global context, leading to improved image quality. Our method achieves competitive results on multiple datasets against the recent deraining methods.
\end{abstract}
\begin{keywords}
Multi-Stage Deraining, Image Restoration, State-Space Models, Frequency-Domain Aware
\end{keywords}
\section{Introduction}
\label{sec:intro}

\begin{figure}[!t]
\centering
\includegraphics[width=0.52\textwidth]{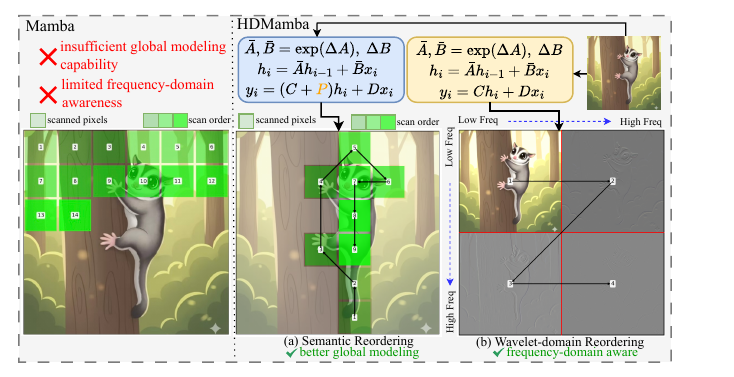}\vspace{-10pt}
\caption{HDMamba provides stronger global modeling and incorporates frequency-domain awareness.}
\label{fig:HDMamba}
\vspace{-12pt}
\end{figure}

Image deraining removes or reduces rain streaks and water droplets from an image to improve the clarity and visibility \cite{10945649,gao2024efficient,10379038}. Adverse weather often produces rain streaks and visibility loss that impair perception. Consequently, robust image deraining is crucial for autonomous driving and surveillance. This task is challenging due to the intricate diversity of rain patterns and difficulty on global consistency with fine-grained recovery.
To address these challenges, recent methods mainly follow two directions, i.e., Convolutional Neural Network (CNN) and Transformer architectures. Representative CNN architectures like UNet employ symmetric encoder-decoder structures with skip connections, enabling multi-scale feature fusion and stable training. While CNN effectively leverages local convolutions for hierarchical feature representation, their limited receptive fields pose challenges for modeling long-range dependencies, particularly on complex rain distributions. This limitation motivates the adoption of global modeling mechanisms ‌such as‌ Transformers in the deraining framework. Transformer methods \cite{zamir2022restormer, liu2024multi, ren2023semi, sun2024haformer} introduce self-attention to model the global context. Although transformers are good at low-frequency and long-range contexts, they are not sensitive to high-frequency rain cues. Beyond transformer-based global attention, state-space models (SSMs) are increasingly employed in the image deraining  task \cite{guo2025mambairv2, wang2025restormamba, li2025ms, zou2024freqmamba}. The structured state-space model \cite{gu2022efficiently} enables long-range dependency modeling with linear complexity. For instance, Mamba's selective scanning \cite{gu2024mamba}  builds correlations among image patches and propagates information across distant regions. However, Mamba models are more responsive to high-frequency information than transformers, they still lack a dedicated frequency-aware component \cite{ren2023semi,yan2024fpgnet}, as illustrated in Fig.~\ref{fig:HDMamba}. To effectively addressing this limitation, our HDMamba module introduces a semantic reordering branch (a) for strong global modeling  and a dedicated wavelet-domain reordering branch (b) to provide frequency-aware.

In this work, we propose Progressive Rain removal with Integrated State-space Modeling (PRISM), which balances the global consistency with fine-grained recovery and frequency-domain fidelity. The PRISM mainly consists of three key stages, i.e., CENet, SFNet and RNet. Specifically, CENet performs preliminary rain removal with HA-UNet, which extracts shallow features and supplies coarse deraining information for the next stage. Then SFNet integrates HA-UNet with HDMamba to jointly model spatial semantics and wavelet domain characteristics. Finally, RNet recovers fine structures via original-resolution subnetwork (ORS). Overall, PRISM adopts a multi-stage architecture, which strengthens high-frequency reconstruction while maintaining global coherence. \noindent Our main contributions are summarized as follows:

\begin{figure}[htb]
\centering
\includegraphics[width=0.49\textwidth]{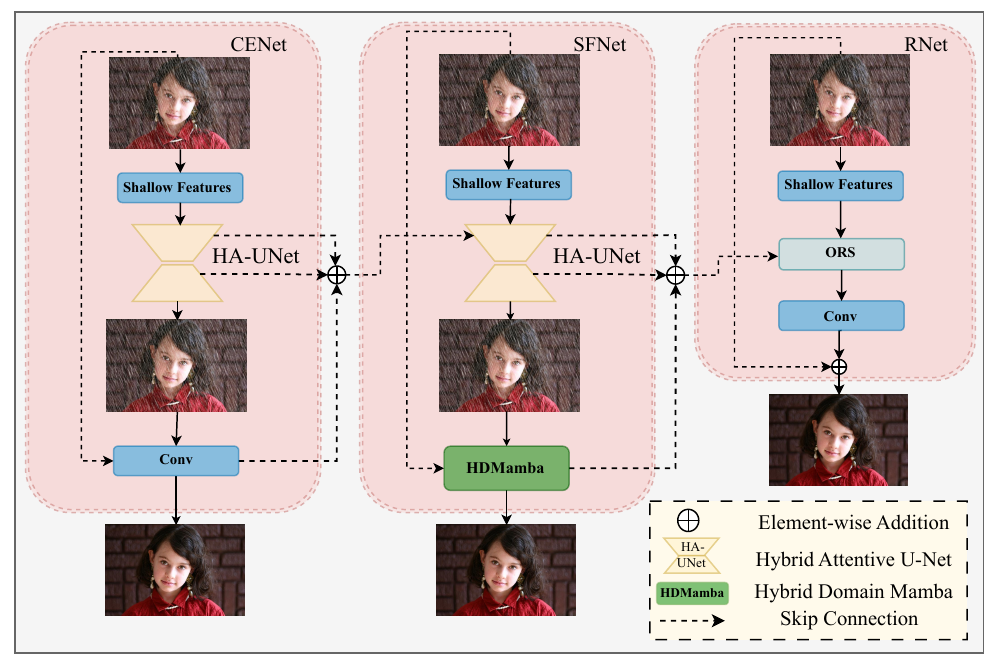}\vspace{-12pt}
\caption{The overall architecture of our proposed PRISM with CENet, SFNet, and RNet to achieve the rain-free image.}
\label{fig:PRISM}
\vspace{-10pt}
\end{figure}
\noindent\textbullet\ We introduce PRISM, a progressive multi-stage deraining framework that enables coarse-to-fine restoration. It not only effectively learns high-frequency rain cues, but also captures global context.   

\noindent\textbullet\ We propose a HA-UNet for the CENet and SFNet stages, which enhances multi-scale feature aggregation by combining Channel Attention (CA) with windowed spatial transformers.  

\noindent\textbullet\ We propose HDMamba that couples state-space sequence modeling with wavelet-based decomposition, to strengthen frequency-aware modeling with efficient global context. 

\noindent\textbullet\ Our experiments on widely used benchmarks demonstrate competitive results. At the same time, our method shows good deraining performance across various rain scenarios.

\section{Method}
\label{sec:method}
As shown in Fig.~\ref{fig:PRISM}, PRISM consists of three key stages: CENet, SFNet, and RNet. CENet and SFNet share a common design with same shallow feature extractor, HA-UNet backbone, and output head. The differences are that CENet uses several convolutional layers for coarse rain-streak removal and SFNet stacks HDMamba modules to fuse wavelet-domain features with global context. And RNet refines the reconstruction with ORS. 

We invent CENet, which extracts shallow features and performs coarse-grained deraining, to model the local spatial context and short-range dependencies. SFNet, which is following and guided by CENet, further refines deraining. The shallow extractor and HA-UNet supply robust representations and multi-scale interactions. The proposed HDMamba achieves global semantic modeling and hybrid domain fusion. It combines semantic reordering for long-range dependencies with a wavelet branch for high-frequency enhancement. An adaptive gating module fuses the spatial and frequency cues, to improve detail recovery and suppress artifacts. RNet merges features and completes the fine-grained reconstruction with ORS. As discussed in prior work \cite{zamir2021multi}, the ORS is effective for fine-grained refinement at the original resolution. 

\subsection{Hybrid Attention UNet}
\begin{figure}[tb]
\centering
\includegraphics[width=0.48\textwidth]{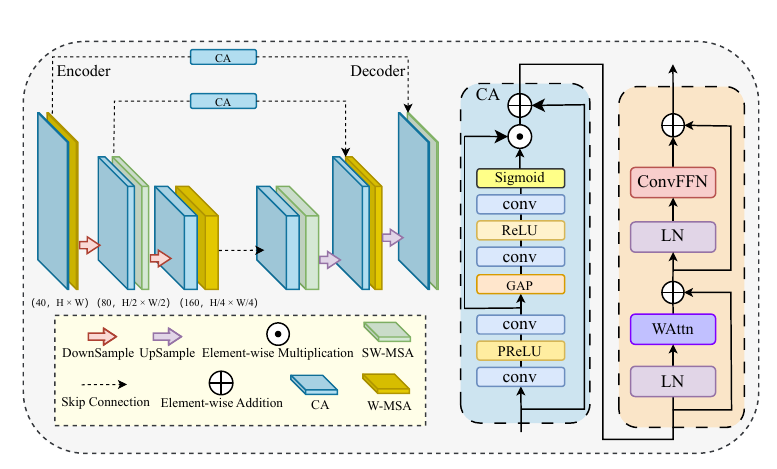}\vspace{-10pt}
\caption{HA‑UNet architecture. A UNet encoder-decoder with hybrid attention and skip connections for feature fusion.}
\label{fig:HA-Unet}
\vspace{-7pt}
\end{figure}

The novel HA-UNet is the core module of our CENet and SFNet, as shown in Fig. \ref{fig:HA-Unet}. UNet has demonstrated strong performance on image deraining in many prior works \cite{zamir2021multi, chen2024bidirectional, fu2023continual, chen2023learning}. In this work, we enhance its expressive power by integrating a joint channel and spatial attention mechanism. In particular, the proposed spatial attention adopts a novel Swin-like windowed self-attention module to capture global spatial context \cite{liu2021swin}. Concretely, HA-UNet employs one basic unit composed of CA, alternating Window Multi-Head Self-Attention (W-MSA) and Shifted Multi-Head Self-Attention (SW-MSA), and one Convolutional Feed-Forward Network (ConvFFN). The specific calculation process of CA is as follows:
\begin{equation}
y=\operatorname{Conv}_{3\times3}\!\big(\phi(\operatorname{Conv}_{3\times3}(x_{\mathrm{in}}))\big),\quad x_{\mathrm{in}} \in \mathbb{R}^{B\times C\times H\times W},
\end{equation}
\begin{equation}
z=\operatorname{GAP}(y),\quad s=\sigma\!\big(\omega_2\,\delta_R(\omega_1 z)\big),
\end{equation}
\begin{equation}
x_{\mathrm{ca}}=x_{\mathrm{in}}+y \odot s,
\end{equation}
where, $x_{\mathrm{in}}\in\mathbb{R}^{B\times C\times H\times W}$ represents the input, $B$ is the batch size, $C$ is the number of channels, $H$ is the image height, and $W$ is the width. The operator $\operatorname{Conv}_{3\times 3}$ uses padding of 1 to preserve spatial resolution and $\phi(\cdot)$ denotes the PReLU activation function. Global Average Pooling (GAP) averages over the $H\times W$ spatial dimensions to get $z\in\mathbb{R}^{B\times C}$. The CA signal $s\in\mathbb{R}^{B\times C}$ is broadcast spatially across $H$ and $W$. The reduction ratio is $r$, with $\omega_1\in\mathbb{R}^{\frac{C}{r}\times C}$ and $\omega_2\in\mathbb{R}^{C\times \frac{C}{r}}$ implemented as $1\times 1$ convolutions or linear layers. Here, $\sigma$ denotes the Sigmoid activation function, and $\delta_R$ denotes the ReLU activation function. In addition, $\odot$ denotes element-wise multiplication.  
\begin{equation}
x_{\mathrm{attn}} = \mathrm{WAttn}\bigl(\mathrm{LN}(\mathrm{CA}(\mathrm{LN}(x_{\mathrm{ca}})))\bigr),
\end{equation}
\begin{equation}
x_{\mathrm{convfnn}} = \mathrm{ConvFNN}\!\big(\mathrm{LN}(x_{\mathrm{attn}})\big).
\end{equation}
‌Here, WAttn is specifically implemented by alternately using W-MSA and SW-MSA, as shown in Fig. \ref{fig:HA-Unet}. LN denotes Layer Normalization. The calculation process of the final component‌ ConvFNN‌ in HA-UNet‌ is as follows:
\begin{equation}
U=\delta_G\!\big(\mathrm{LN}(x_{\mathrm{convfnn}})\,\omega_1+b_1\big),
\end{equation}
\begin{equation}
U'=\mathrm{DWConv}(U),
\end{equation}
\begin{equation}
x_{\mathrm{out}}=U'\,\omega_2+b_2,
\end{equation}
where the $\delta_G$ employs the GELU activation function. We use DWConv to denote a depthwise convolution to preserve spatial size and hidden channels. The ConvFNN projections use $\omega_1\in\mathbb{R}^{C\times C_{\mathrm{ff}}}$ and $\omega_2\in\mathbb{R}^{C_{\mathrm{ff}}\times C}$ with biases $b_1,b_2$, where $C_{\mathrm{ff}}$ denotes the FFN intermediate dimension and ff denotes the feed-forward hidden width. These parameters are not shared with the CA path. The final output of ‌HA-UNet‌ is denoted as ‌$y_{\mathrm{out}}$:
\begin{equation}
y_{\mathrm{out}}=x_{\mathrm{attn}}+x_{\mathrm{out}}.
\end{equation}

\subsection{Hybrid Domain Mamba}
Rain streaks in the image mainly expresses as high-frequency bands. However, the traditional Mamba model cannot achieve frequency-aware modeling and global context aggregation. Specifically, the classic Mamba model achieves token interactions via a discretized state-space recurrence: SSMs map an input sequence $x_i$ to outputs $y_i$ through a latent state $h_i$ updated by a stabilized, discretized linear recurrence. The computational process is detailed below:
\begin{equation}
\label{eq:ase-discretization}
\bar{A},\,\bar{B}=\exp(\Delta A),\,\Delta B,
\end{equation}
\begin{equation}
\label{eq:hdm-state}
h_i=\bar{A}\,h_{i-1}+\bar{B}\,x_i,
\end{equation}
\begin{equation}
\label{eq:hdm-output}
y_i=C\,h_i+D\,x_i,
\end{equation}
where $A$ and $B$ are continuous parameters, while $\bar{A}$ and $\bar{B}$ are discrete parameters, and $\Delta$ is the sampling period, and $exp$ denotes the matrix exponential operation.  Specifically, $\bar{A}$ is the ‌state matrix‌, which governs the system's intrinsic dynamics. Meanwhile, $\bar{B}$ is the ‌input matrix‌, determining how the current input matrix influences and modifies the system state. In addition, $h_i$ denotes the hidden state at the current time step $i$, and $h_{i-1}$ represents the state from the previous time step. However, standard causal state-space models have limited global perception. Each token depends only on predecessors, therefore, long-range interactions decay even with multi-directional scans. 

Accordingly, we propose HDMamba to enhance frequency-aware modeling and global context aggregation for image deraining, as illustrated in Fig.\ref{fig:HDMamba}. This framework employs a semantic reordering Mamba architecture that models spatial-domain semantic reordering. It simultaneously performs wavelet-domain frequency reordering, achieving coordinated cross-domain processing via an adaptive gating fusion mechanism. We follow prior non-causal designs and add a learnable prompt term $P$ to the readout, allowing queries to unscanned positions to enable prompt guided single scan processing \cite{guo2025mambairv2}.


Specifically, in the spatial domain, HDMamba performs semantic reordering. A routing network predicts a semantic class for each spatial position and employs Gumbel-Softmax to realize discrete routing. These features are then rearranged based on semantic affinity. Finally, the prompt guided single scan captures the long-range dependencies within the reordered sequence. This semantic-driven reordering integrates semantically similar regions, improving sequence modeling efficiency and effectiveness. 

Moreover, in the wavelet domain, we perform a standard Discrete Wavelet Transform (DWT) to achieve a two-level decomposition. We then model only the top-level subbands, where the four subbands are arranged as a  $2{\times}2$ quad to form an $H{\times}W$ canvas and flattened into a one-dimensional sequence in a fixed order for a single scan. The input tensor \(I \in \mathbb{R}^{H\times W\times C}\) is decomposed into: one low-frequency approximation map and three oriented high-frequency detail subbands. The detailed computation of DWT proceeds as follows:
\begin{equation}
\label{eq:dwt}
I_{\mathrm{LL}},\, I_{\mathrm{LH}},\, I_{\mathrm{HL}},\, I_{\mathrm{HH}} = \mathrm{DWT}(I),
\end{equation}
where each sub-band corresponds to different directional frequency details. The scanned coefficients are inverse-rearranged and mapped back to the spatial domain via Inverse Discrete Wavelet Transform (IDWT) to yield the wavelet-branch output $X_w$. 

 The two domains are processed in parallel. The spatial semantic reordering branch and the wavelet frequency reordering branch conduct single scans independently to avoid mutual interference. Finally, adaptive gating fuses the two domains, automatically adjusting their relative contributions according to content, so that their advantages can complement each other.
\begin{equation}
\label{eq:hdm-gate-g-fixed}
G=\sigma\!\big([X_s;X_w]\;\omega_g + b_g\big),\quad 
\omega_g\in\mathbb{R}^{2C\times C},\ \ b_g\in\mathbb{R}^{C},
\end{equation}
\begin{equation}
\label{eq:hdm-gate-out}
X_{\text{out}}=G\odot X_s + (1-G)\odot X_w,
\end{equation}
where $G$ is a per-channel gate vector of length $C$ with entries in $(0,1)$. $X_s$ and $X_w$ are the spatial-branch and wavelet-branch features, $[X_s;X_w]$ denotes channel-wise concatenation. $\omega_g$ and $b_g$ are learnable parameters. 
 
\subsection{Loss Function}
We adopt a multi-component loss that combines classical reconstruction terms with a wavelet domain regularization tailored to HDMamba. Following \cite{chen2024bidirectional}, the basic loss includes a Charbonnier loss \cite{zamir2021multi} for robust image reconstruction and an edge loss to preserve sharp structures and fine details. For SFNet, we additionally employ a WaveletLoss to enforce accurate subband reconstruction. Given the prediction \(X_s\) at stage \(s\in\{1,2,3\}\) and ground truth \(Y\), then our global loss function is:
\begin{equation}
\label{eq:loss-global}
\mathcal{L}=\sum_{s=1}^{3}\Big[\mathcal{L}_{\text{char}}(X_s,Y)\;+\;\lambda\,\mathcal{L}_{\text{edge}}(X_s,Y)\;+\;\mu_s\,\mathcal{L}_{\text{wavelet}}(X_s,Y)\Big],
\end{equation}
where, $\mathcal{L}_{char}$ is Charbonnier loss, $\mathcal{L}_{edge}$ is Edge loss, and $\mathcal{L}_{wavelet}$ is Wavelet-domain loss. In our implementation, the coefficient \(\lambda\) in Eq.~\ref{eq:loss-global} balances the Charbonnier and Edge losses and is set to 0.05 as suggested by \cite{zamir2021multi}. For each stage, we set \(\mu_s\in(0,0.05,0)\). The $\mathcal{L}_{\text{char}}$  is described as follows:
\begin{equation}
\label{eq:loss-char}
\mathcal{L}_{\text{char}}(X_s,Y)=\mathbb{E}\!\left[\sqrt{\|X_s-Y\|_2^2+\varepsilon^2}\right],
\end{equation}
where the constant $\varepsilon$ is empirically set to $10^{-3}$ for all the experiments. The $\mathcal{L}_{edge}$ is described as follows:
 \begin{equation}
\label{eq:loss-edge}
\mathcal{L}_{\text{edge}}(X_s,Y)=\sqrt{\|\nabla^2(X_s)-\nabla^2(Y)\|_2^2+\varepsilon^2},
\end{equation}
where $\nabla^2$ denotes the Laplacian operator. The $\mathcal{L}_{wavelet}(X_s,Y)$ is described as follows:
\begin{equation}
\begin{aligned}
\mathcal{L}_{\text{wavelet}}(X_s,Y)
&= \mathcal{L}_1\!\big(LL(X_s),\,LL(Y)\big)\\
&\quad+ \sum_{i=1}^{J}\ \sum_{b\in\{LH,HL,HH\}}
\mathcal{L}_1\!\big(H^{(i)}_{b}(X_s),\,H^{(i)}_{b}(Y)\big),
\end{aligned}
\end{equation}
where, $\mathcal{L}_1$ is the mean of the absolute differences between predictions and targets. $LL_x$ and $LL_y$ denote the level-$J$ low-pass approximation coefficients of the predicted and target images, respectively. $H^{(i)}_{b}$ denotes the $b$-th high-frequency subband at level $i$.


\section{Experiments}
\label{sec:experiments}
In this section, we train PRISM on mixed training deraining datasets and evaluate its performance against representative deraining baselines. The code is publicly available at \url{https://github.com/xuepengze/PRISM}. 
\begin{table*}[t]
\centering
\caption{Quantitative comparison for image deraining on five benchmark datasets.}
\label{tab:Evaluation Metrics}
\resizebox{0.95\textwidth}{!}{
\setlength{\tabcolsep}{5pt}
\renewcommand{\arraystretch}{1.00}
\begin{tabular}{l|cc|cc|cc|cc|cc|cc}
\hline
\multirow{2}{*}{Method} &
\multicolumn{2}{c|}{Test100} &
\multicolumn{2}{c|}{Rain100H} &
\multicolumn{2}{c|}{Rain100L} &
\multicolumn{2}{c|}{Test2800} &
\multicolumn{2}{c|}{Test1200} &
\multicolumn{2}{c}{Average} \\
& PSNR & SSIM & PSNR & SSIM & PSNR & SSIM & PSNR & SSIM & PSNR & SSIM & PSNR & SSIM \\
\hline
DerainNet \cite{fu2017removing}    & 22.77 & 0.810 & 14.92 & 0.592 & 27.03 & 0.884 & 24.31 & 0.861 & 23.38 & 0.835 & 22.48 & 0.796 \\
UMRL \cite{yasarla2019uncertainty}         & 24.41 & 0.829 & 26.01 & 0.832 & 29.18 & 0.923 & 29.97 & 0.905 & 30.55 & 0.910 & 28.02 & 0.880 \\
PReNet \cite{ren2019progressive}       & 24.81 & 0.851 & 26.77 & 0.858 & 32.44 & 0.950 & 31.75 & 0.916 & 31.36 & 0.911 & 29.43 & 0.897 \\
MPRNet \cite{zamir2021multi}       & 30.27 & 0.897 & 30.41 & 0.890 & 36.40 & 0.965 & 33.47 & 0.938 & 32.91 & 0.916 & 32.69 & 0.921 \\
Uformer \cite{wang2022uformer}      & 29.17 & 0.880 & 30.95 & 0.884 & 36.34 & 0.966 & 33.34 & 0.936 & 31.98 & 0.909 & 32.36 & 0.915 \\
Semi-SwinDerain \cite{ren2023semi} & 28.54 & 0.893 & 28.79 & 0.861 & 34.71 & 0.957 & 32.68 & 0.932 & 30.96 & 0.909 & 31.14 & 0.910 \\
DAWN \cite{jiang2023dawn}         & 29.86 & 0.902 & 29.89 & 0.889 & 35.97 & 0.963 & - & - & 32.76 & 0.919 & 32.12 & 0.918 \\
\hline
\textbf{PRISM (Ours)} & 30.29 & 0.900 & 30.06 & 0.889 & 36.88 & 0.966 & 33.73 & 0.939 & 32.56 & 0.913 & 32.70 & 0.921 \\
\hline
\end{tabular}}

\end{table*}

\begin{figure*}[htb]
\centering
\includegraphics[width=7.0in]{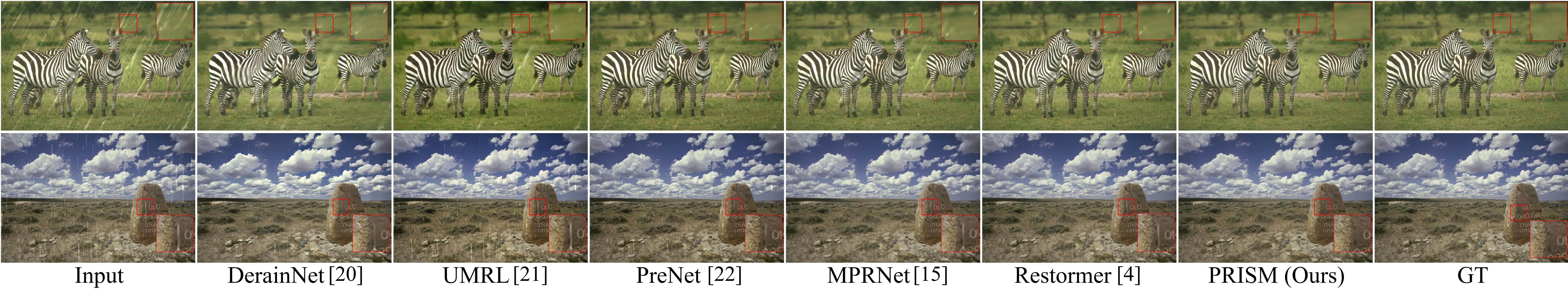}\vspace{-10pt}
\caption{Deraining results on the Rain100L dataset \cite{yang2017deep}. Close-up views highlight deraining details.}
\label{fig:Experience}\vspace{-5pt}
\end{figure*}

\subsection{Datasets}
We follow a mixed training track that integrates classic synthetic deraining datasets. The datasets include Rain800 \cite{li2019single}, Rain100L and Rain100H \cite{yang2017deep}, Rain14000 (DDN-Data) \cite{fu2017removing}, and Rain12 \cite{li2016rain}. In total, this setup yields 13{,}712 paired rainy and clean images. Rain800 has 700 training images and 100 testing images with BSD and UCID backgrounds \cite{li2019single}. Rain100L and Rain100H provide 1{,}800 training pairs and 100 testing images \cite{yang2017deep}. Rain14000 offers 12{,}600 training images with BSD and UCID backgrounds \cite{fu2017removing}. Rain12 has 12 synthesized images, mainly for testing \cite{li2016rain}. This mixed-track strategy improves generalization to varied rain patterns and intensities and better matches real-world deraining scenarios.

\subsection{Evaluation Metrics}
We adopt Peak Signal-to-Noise Ratio (PSNR) and Structural Similarity (SSIM) as objective metrics. PSNR and SSIM reflect pixel-level fidelity and structural similarity, respectively, with higher values indicating better reconstruction and visual quality \cite{you2015adherent,wang2004image}.

\begin{table}[htb]
\centering
\caption{Ablation on the effectiveness of different components. Results are reported on the 100L dataset  \cite{yang2017deep} using PSNR.}
\label{tab:ablation}
\begingroup
\setlength{\tabcolsep}{3pt} 

\begin{tabular}{@{} c c c c @{\hspace{3pt}}|@{\hspace{3pt}} c c c @{}}
\toprule
\multicolumn{4}{c@{\hspace{3pt}}|@{\hspace{3pt}}}{Stage Ablation} & \multicolumn{3}{c}{Module Ablation} \\
CENet & SFNet & RNet & PSNR & HDMamba & HA-UNet & PSNR \\
\midrule
\checkmark &   &   & 34.49 &   & \checkmark & 36.09 \\    
  & \checkmark &   & 33.31 &   &            &       \\    
  &   & \checkmark & 33.33 & \checkmark &   & 34.84 \\    
\midrule
\checkmark & \checkmark & \checkmark & \textbf{36.88} & \checkmark & \checkmark & \textbf{36.88} \\
\bottomrule
\end{tabular}
\endgroup
\vspace{-5pt}
\end{table}

\subsection{Experimental Results}
Following prior works \cite{chen2024bidirectional,guo2025mambairv2,dong2025channel}, we calculate PSNR and SSIM on the Y channel in the YCbCr color space. Across five benchmarks in the mixed training track, PRISM achieves the highest overall average PSNR and SSIM, as shown in Tab.~\ref{tab:Evaluation Metrics}. It obtains better results than the recent Transformer-based Semi-SwinDerain \cite{ren2023semi} and wavelet-based DAWN \cite{jiang2023dawn}, with pronounced gains on Rain100L and Test2800, and comparable performance on Test1200. Then, we further compare PRISM with widely used deraining baselines under the mixed-track setting and present visual results as shown in Fig.~\ref{fig:Experience}. In the first-row zebra example, our model removes slanted rain cues more thoroughly than others, especially in the top-right region, where recent methods leave small raindrop residues. In the second-row stone example, other models tend to misinterpret rain cues as letters on the stone, leading to an erroneous retention of these rain cues. In contrast, PRISM learns to recognize and remove thin and elongated rain streaks, while preserving the integrity of the actual stone letters. These observations indicate robust deraining quality across the rain patterns shown here.

\subsection{Ablation Studies}
\textbf{Number of stages.} We train CENet, SFNet, and RNet as independent single-stage models at the original resolution under the same settings to assess the benefit of a multi-stage design. As shown in Tab.~\ref{tab:ablation}, none of the single-stage variants matches the full three-stage PRISM. RNet further integrates and leverages the information from the preceding CENet and SFNet stages, leading to strong performance. Overall, these results demonstrate that the three-stage fused network is superior to any single-stage counterpart.

\noindent\textbf{Module ablation.} We replace HA-UNet with a standard convolutional UNet of the same depth and width. We further replace HDMamba with a stack of plain convolutional blocks , keeping other settings unchanged. As shown in Tab.~\ref{tab:ablation}, both substitutions degrade the performance, indicating that HA\mbox{-}UNet and HDMamba are effective and complementary.

\section{Conclusion}
We propose PRISM, an efficient three-stage progressive deraining network for fine-grained recovery with global consistency. The PRISM mainly consists of three key stages, i.e., CENet, SFNet and RNet. Specifically, CENet conducts initial rain removal using HA-UNet. Then SFNet integrates HA-UNet with HDMamba to extract shallow features and provide coarse deraining. Finally, RNet recovers fine structures via ORS. The integration‌ of HA-UNet and HDMamba ‌facilitates‌ balanced preservation of ‌local details‌ ‌and‌ global consistency ‌throughout the deraining process. Moreover, our approach enhances restoration in local regions and captures critical high-frequency rain-streak clues to remove residual artifacts. PRISM achieves competitively quantitative results and stable visual quality on standard synthetic benchmarks, with restored images showing nearly no residual streaks.



\bibliographystyle{IEEEbib}
\bibliography{refs}

\end{document}